\documentclass{article}
\usepackage{iclr2021_conference,times}

\usepackage[utf8]{inputenc} 
\usepackage[T1]{fontenc}    
\usepackage{hyperref}       
\usepackage{url}            
\usepackage{booktabs}       
\usepackage{amsfonts}       
\usepackage{amsmath}
\usepackage{nicefrac}       
\usepackage{microtype}      
\usepackage{todonotes}
\usepackage[ruled,vlined,linesnumbered]{algorithm2e}
\usepackage{subcaption}
\usepackage{array}
\usepackage{arydshln}
\usepackage[group-separator={,}]{siunitx}
\usepackage{dutchcal}
\usepackage{wrapfig}
\usepackage{float}

\usepackage{natbib}
\title{Temporal Graph Networks for Deep Learning on Dynamic Graphs}

\author{%
  Emanuele Rossi\thanks{erossi@twitter.com} \\
  Twitter \\
   \And
   Ben Chamberlain \\
   Twitter \\
   \And
   Fabrizio Frasca \\
   Twitter \\
   \AND
   Davide Eynard \\
   Twitter \\
   \And
   Federico Monti \\
   Twitter \\
   \And
   Michael Bronstein \\
   Twitter \\
   \\
}

\iclrfinalcopy 
\begin{document}

\maketitle

\begin{abstract}
Graph Neural Networks (GNNs) have recently become increasingly popular due to their ability to learn complex systems of relations or interactions. These arise in a broad spectrum of problems ranging from biology and particle physics to social networks and recommendation systems. Despite the plethora of different models for deep learning on graphs, few approaches have been proposed for dealing with graphs that are dynamic in nature (e.g. evolving features or connectivity over time). 
We present Temporal Graph Networks (TGNs), a generic, efficient framework for deep learning on dynamic graphs represented as sequences of timed events. Thanks to a novel combination of memory modules and graph-based operators, TGNs significantly outperform previous approaches while being more computationally efficient. 
We furthermore show that several previous models for learning on dynamic graphs can be cast as specific instances of our framework. 
We perform a detailed ablation study of different components of our framework and devise the best configuration that achieves state-of-the-art performance on several transductive and inductive prediction tasks for dynamic graphs. 

\end{abstract}

\section{Introduction}

In the past few years, graph representation learning \citep{bronstein2017geometric,hamilton2017representation,battaglia2018relational} has produced a sequence of successes, gaining increasing popularity in machine learning. 
Graphs are ubiquitously used as models for systems of relations and interactions in many fields \citep{battaglia2016interaction,qi2018learning,Monti2016GeometricDL,choma2018graph,NIPS2015_5954,pmlr-v70-gilmer17a,parisot2018disease,rossi2019ncrna}, in particular, social sciences \citep{pinsage,monti2019fake,rossi2020sign} and biology \citep{zitnik2018modeling,veselkov2019hyperfoods,gainza2019deciphering}. Learning on such data is possible using graph neural networks (GNNs) 
\citep{graphsage}
that typically operate by a message passing mechanism \citep{battaglia2018relational} aggregating information in a neighborhood of a node and create node embeddings that are then used for node classification \citep{Monti2016GeometricDL, DBLP:conf/iclr/VelickovicCCRLB18, Kipf:2016tc}, graph classification~\citep{pmlr-v70-gilmer17a}, or edge prediction \citep{zhang2018link} tasks.

The majority of methods for deep learning on graphs assume that the underlying graph is {\em static}. However, most real-life systems of interactions such as social networks or biological interactomes are {\em dynamic}. While it is often possible to apply static graph deep learning models~\citep{10.5555/1241540.1241551} to dynamic graphs by ignoring the temporal evolution, this has been shown to be sub-optimal \citep{Xu2020Inductive}, and in some cases, it is the dynamic structure that contains crucial insights about the system. 
Learning on dynamic graphs is relatively recent, and most works are limited to the setting of discrete-time dynamic graphs represented as a sequence of snapshots of the graph~\citep{10.5555/1241540.1241551, dunlavy2011temporal, yu20193d, sankar2020dysat, pareja2019evolvegcn, yu2018netwalk}. Such approaches are unsuitable for interesting real world settings such as social networks, where dynamic graphs are continuous (i.e. edges can appear at any time) and evolving (i.e. new nodes join the graph continuously). Only recently, several approaches have been proposed that support the continuous-time scenario~\citep{Xu2020Inductive, trivedi2018dyrep, 10.1145/3292500.3330895, ma2018streaming, 8622109, bastas2019evolve2vec}. %

\paragraph{Contributions.}
In this paper, we first propose the generic inductive framework of Temporal Graph Networks (TGNs) operating on continuous-time dynamic graphs represented as a sequence of events, and show that many previous methods are specific instances of TGNs. 
Second, we propose a novel training strategy allowing the model to learn from the sequentiality of the data while maintaining highly efficient parallel processing. 
Third, we perform a detailed ablation study of different components of our framework and analyze the tradeoff between speed and accuracy. 
Finally, we show state-of-the-art performance on multiple tasks and datasets in both transductive and inductive settings, while being much faster than previous methods. 

\section{Background}

\paragraph{Deep learning on static graphs.}
A static graph $\mathcal{G}=(\mathcal{V}, \mathcal{E})$ comprises nodes $\mathcal{V} = \{1, \hdots, n\}$ and edges $\mathcal{E} \subseteq \mathcal{V} \times \mathcal{V}$, which are endowed with features, denoted by $\mathbf{v}_i$ and $\mathbf{e}_{ij}$ for all $i,j = 1,\hdots, n$, respectively. 
A typical {\em graph neural network} (GNN) creates an {\em embedding} $\mathbf{z}_i$ of the nodes by learning a local aggregation rule of the form\vspace{-1mm}
$$
\mathbf{z}_i = \sum_{j \in \mathcal{N}_i} h(\mathbf{m}_{ij}, \mathbf{v}_i
) \quad\quad\quad \mathbf{m}_{ij} = \mathrm{msg}(\mathbf{v}_i,\mathbf{v}_j,\mathbf{e}_{ij}),
\vspace{-1mm}
$$
which is interpreted as message passing from the neighbors $j$ of $i$. 
Here, $\mathcal{N}_i = \{ j : (i,j) \in \mathcal{E} \}$ denotes the neighborhood of node $i$ and $\mathrm{msg}$ and $h$ are learnable functions.


\paragraph{Dynamic Graphs.}
There exist two main models for dynamic graphs. 
 \textit{Discrete-time dynamic graph}s (DTDG) are sequences of static graph snapshots taken at intervals in time. 
 {\em Continuos-time dynamic graph}s (CTDG) are more general and can be represented as timed lists of events, which may include edge addition or deletion, node addition or deletion and node or edge feature transformations. 

Our temporal (multi-)graph is 
modeled as a sequence of time-stamped {\em events} $\mathcal{G} = \{ x(t_1), x(t_2), \hdots \}$, representing addition or change of a node or interaction between a pair of nodes at times $0\leq t_1 \leq t_2 \leq \hdots$. 
An event $x(t)$ can be of two types: 
1) A {\bf node-wise event} is represented by $\mathbf{v}_i(t)$, where $i$ denotes the index of the node and $\mathbf{v}$ is the vector attribute associated with the event. 
If the index $i$ has not been seen before, the event creates node $i$ (with the given features), otherwise it updates the features.    
2) An {\bf interaction event} between nodes $i$ and $j$ is represented by a (directed) {\em temporal edge} $\mathbf{e}_{ij}(t)$ (there might be more than one edge between a pair of nodes, so technically this is a multigraph).  
We denote by 
$\mathcal{V}(T) = \{ i \, : \, \exists\mathbf{v}_{i}(t) \in \mathcal{G}, t\in T \}$
and 
$\mathcal{E}(T) = \{ (i,j) \, : \, \exists\mathbf{e}_{ij}(t) \in \mathcal{G}, t\in T \}$ the temporal set of vertices and edges, respectively, 
and by $\mathcal{N}_i(T) = \{ j \, : \, (i,j) \in  \mathcal{E}(T)\}$ the neighborhood of node $i$ in time interval $T$. $\mathcal{N}^k_i(T)$ denotes the $k$-hop neighborhood. 
A {\em snapshot} of the temporal graph $\mathcal{G}$ at time $t$ is the (multi-)graph $\mathcal{G}(t) = ( \mathcal{V}[0,t], \mathcal{E}[0,t])$ with $n(t)$ nodes. 
{\bf Deletion events} are discussed in Appendix~\ref{sec:deletion_events}. 


\section{Temporal Graph Networks}

Following the terminology in~\citep{dynamic-graphs-survey}, a neural model for dynamic graphs can be regarded as an encoder-decoder pair, where an encoder is a function that maps from a dynamic graph to node embeddings, and a decoder takes as input  one or more node embeddings and makes a task-specific prediction e.g. node classification or edge prediction.
The key contribution of this paper is a novel Temporal Graph Network (TGN) encoder applied on a continuous-time dynamic graph represented as a sequence of time-stamped events and producing, for each time $t$, the embedding of the graph nodes $\mathbf{Z}\left(t) = (\mathbf{z}_1(t), \hdots, \mathbf{z}_{n(t)}(t)\right)$.

\subsection{Core modules} \label{sec:core_modules}

\begin{figure}\vspace{-4mm}
    \centering
    \includegraphics[width=.85\textwidth]{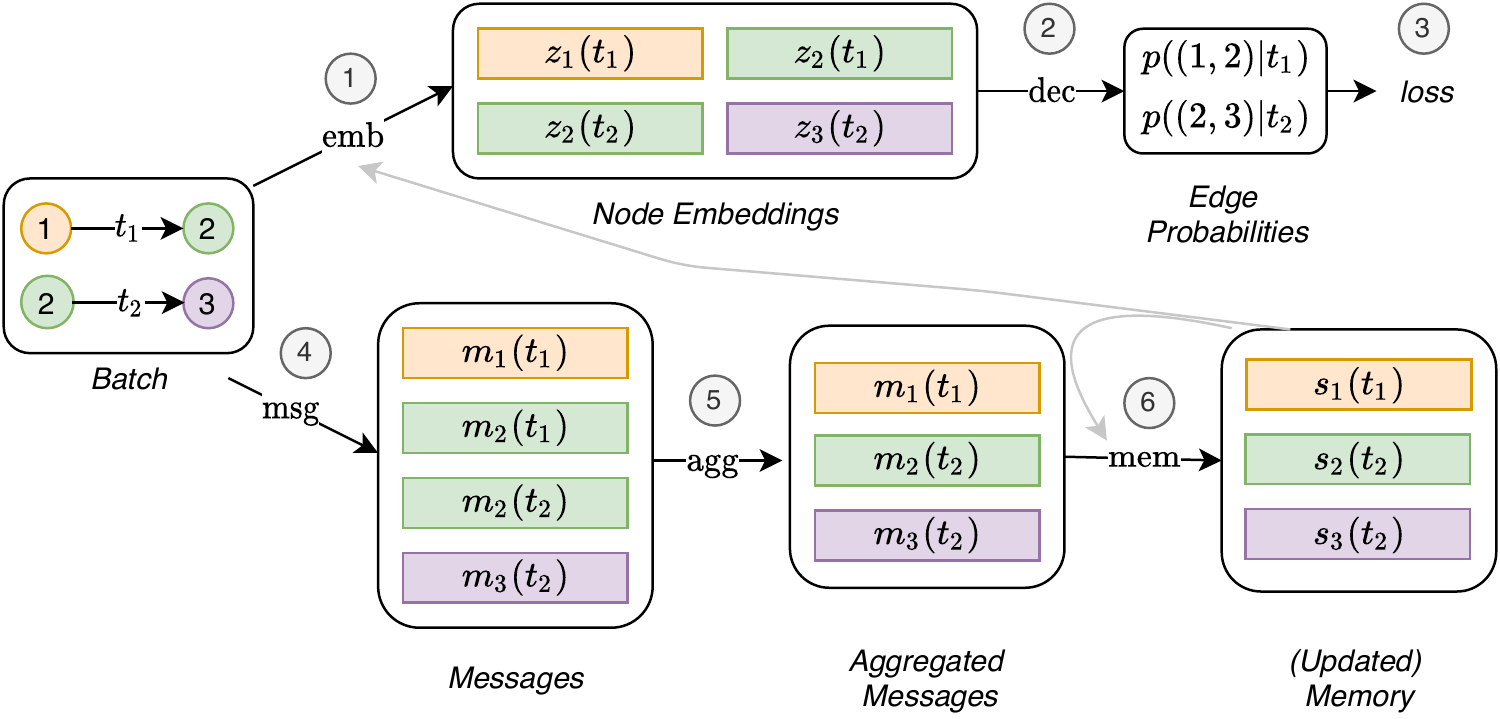}\vspace{-1mm}
    \caption{Computations performed by TGN on a batch of time-stamped interactions. {\em Top:} embeddings are produced by the embedding module using the temporal graph and the node’s memory (1). The embeddings are then used to predict the batch interactions and compute the loss (2, 3). {\em Bottom:} these same interactions are used to update the memory (4, 5, 6). This is a simplified flow of operations which would prevent the training of all the modules in the bottom as they would not receiving a gradient. Section \ref{sec:training} explains how to change the flow of operations to solve this problem and figure \ref{fig:diagram_training} shows the complete diagram. }\vspace{-2mm}
    \label{fig:diagram}
\end{figure}

\paragraph*{Memory.}
The memory (state) of the model at time $t$ consists of a vector $\mathbf{s}_i(t)$ for each node $i$ the model has seen so far. The memory of a node is updated after an event (e.g. interaction with another node or node-wise change), and its purpose is to represent the node's history in a compressed format. Thanks to this specific module, TGNs have the capability to memorize long term dependencies for each node in the graph. When a new node is encountered, its memory is initialized as the zero vector, and it is then updated for each event involving the node, even after the model has finished training.
While a global (graph-wise) memory can also be added to the model to track the evolution of the entire network, we leave this as future work.

\paragraph*{Message Function.}

For each event involving node $i$, a message is computed to update $i$'s memory. 
In the case of an interaction event $\mathbf{e}_{ij}(t)$ between source node $i$ and target node $j$ at time $t$,  
two messages can be computed:\vspace{-1mm}
\begin{eqnarray}
    \mathbf{m}_i(t) = \mathrm{msg_s}\left(\mathbf{s}_i(t^-), \mathbf{s}_j(t^-), \Delta t, \mathbf{e}_{ij}(t)\right), \quad\quad
    \mathbf{m}_j(t) = \mathrm{msg_d}\left(\mathbf{s}_j(t^-), \mathbf{s}_i(t^-), \Delta t, \mathbf{e}_{ij}(t)\right)\vspace{-2mm}
\end{eqnarray}
Similarly, in case of a node-wise event $\mathbf{v}_i(t)$, a single message can be computed for the node involved in the event: \vspace{-2mm}
\begin{eqnarray}
    \mathbf{m}_i(t) = \mathrm{msg_n}\left(\mathbf{s}_i(t^-), t, \mathbf{v}_{i}(t)\right).\vspace{-1mm} 
\end{eqnarray}
Here, $\mathbf{s}_i(t^-)$ is the memory of node $i$ just before time $t$ (i.e., from the time of the previous interaction involving $i$) 
and $\mathrm{msg_s}, \mathrm{msg_d}$ and $\mathrm{msg_n}$ are learnable message functions, e.g. MLPs. 
In all experiments, we choose the message function as {\em identity} (id), which is simply the concatenation of the inputs, for the sake of simplicity. Deletion events are also supported by the framework and are presented in Appendix~\ref{sec:deletion_events}. 
A more complex message function that involves additional aggregation from the neighbours of nodes $i$ and $j$ is also possible and is left for future study.

\paragraph*{Message Aggregator.}
Resorting to batch processing for efficiency reasons may lead to multiple events involving the same node $i$ in the same batch. As each event generates a message in our formulation, we use a mechanism to aggregate messages $\mathbf{m}_i(t_1), \hdots, \mathbf{m}_i(t_b)$ for $t_1, \hdots, t_b \leq t$,
\begin{eqnarray}
\bar{\mathbf{m}}_i(t) = \mathrm{agg}\left(\mathbf{m}_i(t_1), \hdots, \mathbf{m}_i(t_b)\right).  
\end{eqnarray}
Here, $\mathrm{agg}$ is an aggregation function.  
While multiple choices can be considered for implementing this module (e.g. RNNs or attention w.r.t. the node memory), for the sake of simplicity we considered two efficient non-learnable solutions in our experiments: {\em most recent message} (keep only most recent message for a given node) and {\em mean message} (average all messages for a given node). We leave learnable aggregation as a future research direction.

\paragraph*{Memory Updater.}
As previously mentioned, the memory of a node is updated upon each event involving the node itself: \vspace{-2mm}
\begin{eqnarray}
    \mathbf{s}_i(t) = \mathrm{mem}\left(\bar{\mathbf{m}}_i(t), \mathbf{s}_i(t^-)\right). 
\end{eqnarray}
For interaction events involving two nodes $i$ and $j$, the memories of both nodes are updated after the event has happened. For node-wise events, only the memory of the related node is updated.
Here, $\mathrm{mem}$ is a learnable memory update function, e.g. a recurrent neural network such as LSTM~\citep{10.1162/neco.1997.9.8.1735} or GRU~\citep{cho-etal-2014-learning}. 


\paragraph*{Embedding.}

The embedding module is used to generate the temporal embedding $\mathbf{z}_i(t)$ of node $i$ at any time $t$. 
The main goal of the embedding module is to avoid the so-called memory staleness problem~\citep{dynamic-graphs-survey}. Since the memory of a node $i$ is updated only when the node is involved in an event, it might happen that, in the absence of events for a long time (e.g. a social network user who stops using the platform for some time before becoming active again), $i$'s memory becomes stale.
While multiple implementations of the embedding module are possible, we use the form:  
\begin{equation} 
    \mathbf{z}_i(t) = \mathrm{emb}(i, t) = \sum_{j \in \mathcal{N}^k_i([0, t]) } h\left(\mathbf{s}_i(t), \mathbf{s}_j(t), \mathbf{e}_{ij}, \mathbf{v}_i(t), \mathbf{v}_j(t)\right), \nonumber
\end{equation}
where $h$ is a learnable function. 
This includes many different formulations as particular cases: 

\noindent {\em Identity} (id): $\mathrm{emb}(i, t) = \mathbf{s}_i(t)$, which uses the memory directly as the node embedding.

\noindent  {\em Time projection} (time): $\mathrm{emb}(i, t) = (1 + \Delta t \, \mathbf{w}) \circ \mathbf{s}_i(t)$, where $\mathbf{w}$ are learnable parameters, $\Delta t$ is the time since the last interaction, and $\circ$ denotes element-wise vector product. This version of the embedding method was used in Jodie~\citep{10.1145/3292500.3330895}.  

\noindent {\em Temporal Graph Attention} (attn): A series of $L$ graph attention layers compute $i$'s embedding by aggregating information from its $L$-hop temporal neighborhood.

The input to the $l$-th layer is $i$'s representation $\mathbf{h}_i^{(l-1)}(t)$, the current timestamp $t$, $i$'s neighborhood representation $\{\mathbf{h}_1^{(l-1)}(t), \ldots, \mathbf{h}_N^{(l-1)}(t)\}$ together with timestamps $t_1, \ldots, t_N$ and features $\mathbf{e}_{i1}(t_1), \ldots, \mathbf{e}_{iN}(t_N)$ for each of the considered interactions which form an edge in $i$'s temporal neighborhood: 
\begin{eqnarray}
    \mathbf{h}_i^{(l)}(t) &=&\mathrm{MLP}^{(l)}(\mathbf{h}_i^{(l-1)}(t) \, \|\, \tilde{\mathbf{h}}_i^{(l)}(t)),\\
    \tilde{\mathbf{h}}_i^{(l)}(t) &=& \mathrm{MultiHeadAttention}^{(l)}(\mathbf{q}^{(l)}(t), \mathbf{K}^{(l)}(t), \mathbf{V}^{(l)}(t)), \\
    \mathbf{q}^{(l)}(t) &=& \mathbf{h}_i^{(l-1)}(t) \, \| \, \boldsymbol{\phi}(0),\\
    \mathbf{K}^{(l)}(t) &=& \mathbf{V}^{(l)}(t) = \mathbf{C}^{(l)}(t), \\
    \mathbf{C}^{(l)}(t) &=& [\mathbf{h}^{(l-1)}_1(t) \,\|\, \mathbf{e}_{i1}(t_1) \,\|\, \boldsymbol{\phi}(t - t_{1}),\, \ldots,\, 
    \mathbf{h}^{(l-1)}_N(t)\, \|\, \mathbf{e}_{iN}(t_N)\, \|\, \boldsymbol{\phi}(t - t_{N})].
\end{eqnarray}

 Here,  $\boldsymbol{\phi}(\cdot)$ represents a generic time encoding \citep{Xu2020Inductive}, $\|$ is the concatenation operator and $\mathbf{z}_i(t) = \mathrm{emb}(i, t) = \mathbf{h}_i^{(L)}(t)$. Each layer amounts to performing multi-head-attention~\citep{NIPS2017_7181} where the query ($\mathbf{q}^{(l)}(t)$) is a reference node (i.e. the target node or one of its $L-1$-hop neighbors), and the keys $\mathbf{K}^{(l)}(t)$ and values $\mathbf{V}^{(l)}(t)$ are its neighbors. Finally, an MLP is used to combine the reference node representation with the aggregated information. 
 Differently from the original formulation of this layer (firstly proposed in TGAT~\citep{Xu2020Inductive}) where no node-wise temporal features were used, in our case the input representation of each node $\mathbf{h}^{(0)}_j(t) = \mathbf{s}_j(t) + \mathbf{v}_j(t)$ and as such it allows the model to exploit both the current memory $\mathbf{s}_j(t)$ and the temporal node features $\mathbf{v}_j(t)$.

\noindent {\em Temporal Graph Sum} (sum):
 A simpler and faster aggregation over the graph:
\begin{eqnarray}
    \mathbf{h}^{(l)}_i(t) &=& \mathbf{W}_2^{(l)}(\mathbf{h}^{(l-1)}_i(t) \, \| \, \tilde{\mathbf{h}}^{(l)}_i(t)),\\
    \tilde{\mathbf{h}}^{(l)}_i(t) &=& \mathrm{ReLu} (\sum_{j  \in \mathcal{N}_i([0, t]) } \mathbf{W}_1^{(l)}(\mathbf{h}^{(l-1)}_j(t) \, \|\, \mathbf{e}_{ij} \, \|\, \boldsymbol{\phi}(t - t_{j}))).
\end{eqnarray}
Here as well, $\boldsymbol{\phi}(\cdot)$ is a time encoding and $\mathbf{z}_i(t) = \mathrm{emb}(i, t) = \mathbf{h}_i^{(L)}(t)$. In the experiment, both for the \textit{Temporal Graph Attention} and for the \textit{Temporal Graph Sum} modules we use the time encoding presented in Time2Vec~\citep{DBLP:journals/corr/abs-1907-05321} and used in TGAT~\citep{Xu2020Inductive}.

The graph embedding modules mitigate the staleness problem by aggregating information from a node's neighbors memory. When a node has been inactive for a while, it is likely that some of its neighbours have been recently active, and by aggregating their memories, TGN can compute an up-to-date embedding for the node. The temporal graph attention is additionally able to select which neighbors are more important based on both features and timing information.

\subsection{Training} \label{sec:training}

TGN can be trained for a variety of tasks such as edge prediction (self-supervised) or node classification (semi-supervised). We use  link prediction  as an example: provided a list of time ordered interactions, the goal is to predict future interactions from those observed in the past. Figure \ref{fig:diagram} shows the computations performed by TGN on a batch of training data. 

\begin{figure}\vspace{-6mm}
    \centering
    \includegraphics[width=.85\textwidth]{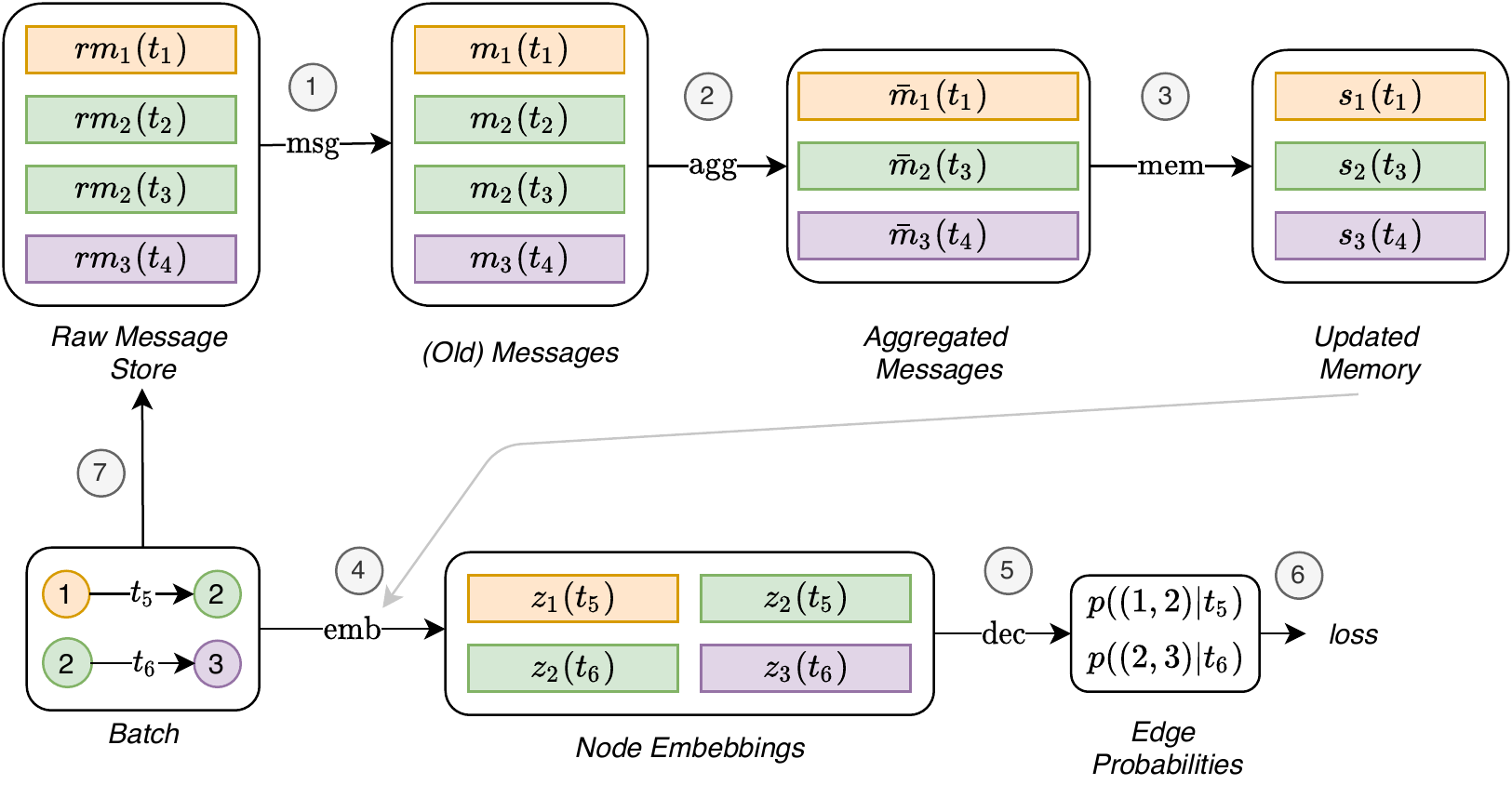}\vspace{-2mm}
    \caption{Flow of operations of TGN used to train the memory-related modules. \textit{Raw Message Store} stores the necessary raw information to compute messages, i.e. the input to the message functions, which we call raw messages, for interactions which have been processed by the model in the past. This allows the model to delay the memory update brought by an interaction to later batches. At first, the memory is updated using messages computed from raw messages stored in previous batches (1, 2, 3). The embeddings can then be computed using the just updated memory (grey link) (4). By doing this, the computation of the memory-related modules directly influences the loss (5, 6), and they receive a gradient. Finally, the raw messages for this batch interactions are stored in the raw message store (6) to be used in future batches.}\vspace{-5mm}
    \label{fig:diagram_training}
\end{figure}

The complexity in our training strategy relates to the memory-related modules (\textit{Message function}, \textit{Message aggregator}, and \textit{Memory updater}) because they do not directly influence the loss and therefore do not receive a gradient. To solve this problem, the memory must be updated before predicting the batch interactions. However, updating the memory with an interaction $\mathbf{e}_{ij}(t)$ before using the model to predict that same interaction, causes information leakage. 
To avoid the issue, when processing a batch, we update the memory with messages coming from previous batches (which are stored in the \textit{Raw Message Store}), and then predict the interactions. Figure \ref{fig:diagram_training} shows the training flow for the memory-related modules. Pseudocode for the training procedure is presented in Appendix~\ref{sec:pseudocode}. 

More formally, at any time $t$, the Raw Message Store contains (at most) one raw message $rm_i$ for each node $i$\footnote{The Raw Message Store does not contain a message for $i$ only if $i$ has never been involved in an event in the past.}, generated from the last interaction involving $i$ before time $t$. When the model processes the next interactions involving $i$, its memory is updated using $rm_i$ (arrows 1, 2, 3 in Figure~\ref{fig:diagram_training}), then the updated memory is used to compute the node's embedding and the batch loss (arrows 4, 5, 6). Finally, the raw messages for the new interaction are stored in the raw message store (arrows 7).
%
It is also worth noticing that all predictions in a given batch have access to the same state of the memory. While from the perspective of the first interaction in the batch the memory is up-to-date (since it contains information about all previous interactions in the graph), from the perspective of the last interaction in the batch the same memory is out-of-date, since it lacks information about previous interactions in the same batch. This disincentives the use of a big batch size (in the extreme case where the batch size is a big as the dataset, all predictions would be made using the initial zero memory). We found a batch size of $200$ to be a good trade-off between speed and update granularity.

\section{Related Work}


Early models for learning on dynamic graphs focused on DTDGs. Such approaches either aggregate graph snapshots and then apply static methods~\citep{10.5555/1241540.1241551, Hisano_2018, sharan2008temporal, ibrahim2015link, ahmed2016efficient, ahmed2016sampling}, assemble snapshots into tensors and factorize~\citep{dunlavy2011temporal, yu2017link, ma2019embedding}, or encode each snapshot to produce a series of embeddings. In the latter case, the embeddings are either aggregated by taking a weighted sum~\citep{yao2016link, zhu2012hybrid}, fit to time series models~\citep{huang2009time, gunecs2016link, da2012time, moradabadi2017novel}, used as components in RNNs~\citep{Seo_2018, NARAYAN2018433, MANESSI2020107000, yu20193d, chen2018gclstm, sankar2020dysat, pareja2019evolvegcn}, or learned by imposing a smoothness constraint over time~\citep{kim2009particle, gupta2011evolutionary, yao2016link, zhu2017next, zhou2018dynamic, ijcai2019-640, journals/corr/abs-1805-11273, fard2019relationship, pei2016node}. Another line of work encodes DTDGs by first performing random walks on an initial snapshot and then modifying the walk behaviour for subsequent snapshots~\citep{mahdavi2018dynnode2vec,du2018dynamic, xin2016adaptive, 8508272, yu2018netwalk}. 
Spatio-temporal graphs (considered by \cite{zhang2018gaan, li2018diffusion} for traffic forecasting) are specific cases of dynamic graphs where the topology of the graph is fixed.

CTDGs have been addressed only recently. Several approaches use random walk models \citep{10.1145/3184558.3191526, 8622109, bastas2019evolve2vec} incorporating continuous time through constraints on transition probabilities. Sequence-based approaches for CTDGs~\citep{10.1145/3292500.3330895, 10.5555/3305890.3306039, trivedi2018dyrep, ma2018streaming} use RNNs to update representations of the source and destination node each time a new edge appears. Other recent works have focused on dynamic knowledge graphs ~\citep{goel2019diachronic, xu2019temporal, dasgupta-etal-2018-hyte, garcia-duran-etal-2018-learning}. 
Many architectures for continuous-time dynamic graphs are based on a node-wise memory updated by an RNN when new interactions appear. Yet, they lack a GNN-like aggregation from a node's neighbors when computing its embedding, which makes them susceptible to the staleness problem (i.e. a node embedding becoming out of date) while at the same time also limiting their expressive power.

Most recent CTDGs learning models can be interpreted as specific cases of our framework (see Table \ref{tab:models}). For example, 
Jodie~\citep{10.1145/3292500.3330895} uses the time projection embedding module $\mathrm{emb}(i, t) = (1 + \Delta t \mathbf{w}) \circ \mathbf{s}_i(t)$. 
TGAT~\citep{Xu2020Inductive} is a specific case of TGN when the memory and its related modules are missing, and graph attention is used as the Embedding module.  
DyRep~\citep{trivedi2018dyrep} computes messages using graph attention on the destination node neighbors. 
%
%
Finally, we note that TGN generalizes the Graph Networks (GN) model \citep{battaglia2018relational} for static graphs (with the exception of the global block that we omitted from our model for the sake of simplicity), and thus the majority of existing graph message passing-type architectures.


\begin{table}[h!]
  \caption{Previous models for deep learning on continuous-time dynamic graphs are specific case of our TGN framework. Shown are multiple variants of TGN used in our ablation studies. {\em method} ($l$,$n$) refers to graph convolution using $l$ layers and $n$ neighbors.  $^\dagger$uses t-batches. $^*$ uses uniform sampling of neighbors, while the default is sampling the most recent neighbors. $^\ddagger$message aggregation not explained in the paper. $^\|$ uses a summary of the destination node neighborhood (obtained through graph attention) as additional input to the message function.
  } 

  \label{tab:models}
  \centering
  \begin{tabular}{lccccc}
    \toprule
              & Mem.          & Mem. Updater           & Embedding             & Mess. Agg.                               &  Mess. Func.    \\
    \midrule
    Jodie          & node     & RNN                    & time                  & \hspace{1.5mm}---$^\dagger$\hspace{0mm}  & id              \\ 
    TGAT           & ---      & ---                    & attn (2l, 20n)$^*$    & ---                                      & ---             \\
    DyRep          & node     & RNN                    & id                    & \hspace{1.5mm}---$^\ddagger$\hspace{0mm} & \hspace{1.5mm}attn$^\|$\hspace{0mm}        \\ \hline 
    TGN-attn       & node     & GRU                    & attn (1l, 10n)        & last                                     & id              \\
    TGN-2l         & node     & GRU                    & attn (2l, 10n)        & last                                     & id              \\
    TGN-no-mem     & ---      & ---                    & attn (1l, 10n)        & ---                                      & ---             \\
    TGN-time       & node     & GRU                    & time                  & last                                     & id              \\
    TGN-id         & node     & GRU                    & id                    & last                                     & id              \\
    TGN-sum        & node     & GRU                    & sum  (1l, 10n)        & last                                     & id              \\
    TGN-mean       & node     & GRU                    & attn (1l, 10n)        & mean                                     & id              \\
    \bottomrule
  \end{tabular}\vspace{-2mm}
\end{table}

\section{Experiments} \label{sec:experiments}


\paragraph{Datasets.} 
We use three datasets in our experiments: Wikipedia, Reddit ~\citep{10.1145/3292500.3330895}, and Twitter, which are described in detail in Appendix~\ref{sec:datasets}.
%
Our experimental setup closely follows \citep{Xu2020Inductive} and focuses on the tasks of future edge (`link') prediction and dynamic node classification. 
In future edge prediction, the goal is to predict the probability of an edge occurring between two nodes at a given time. Our encoder is combined with a simple MLP decoder mapping from the concatenation of two node embeddings to the probability of the edge. We study both the transductive and inductive settings. In the transductive task, we predict future links of the nodes observed during training, whereas in the inductive tasks we predict future links of nodes never observed before. 
For node classification, the transductive setting is used.
For all tasks and datasets we perform the same 70\%-15\%-15\% chronological split as in \citet{Xu2020Inductive}. 
All the results were averaged over 10 runs. 
Hyperparameters and additional details can be found in Appendix~\ref{sec:additional_exp_setting}.


\paragraph{Baselines.} Our strong baselines are state-of-the-art approaches for continuous time dynamic graphs (CTDNE~\citep{10.1145/3184558.3191526}, Jodie~\citep{10.1145/3292500.3330895}, DyRep~\citep{trivedi2018dyrep} and TGAT~\citep{Xu2020Inductive}) as well as state-of-the-art models for static graphs (GAE~\citep{kipf2016variational}, VGAE~\citep{kipf2016variational}, DeepWalk~\citep{10.1145/2623330.2623732},  Node2Vec~\citep{10.1145/2939672.2939754}, GAT \citep{DBLP:conf/iclr/VelickovicCCRLB18} and GraphSAGE \citep{hamilton2017representation}).

\begin{table}
  \caption{Average Precision (\%) for future edge prediction task in transductive and inductive settings. {\bf \bf \color{red} First}, {\bf \bf \color{violet} Second}, {\bf Third} best performing method. $^*$Static graph method.  $^\dagger$Does not support inductive.}
  \label{tab:edge-prediction}
  \centering
  \hspace*{-0.82cm}
  \begin{tabular}{lcccccc}
    \toprule
                     & \multicolumn{2}{c}{Wikipedia}                    & \multicolumn{2}{c}{Reddit}                        & \multicolumn{2}{c}{Twitter}                       \\
    \cmidrule(r){2-7} 
                     & Transductive            & Inductive              & Transductive            & Inductive               & Transductive            & Inductive               \\
    \midrule
    GAE$^*$          & $91.44\pm0.1$           & $^\dagger$             & $93.23\pm0.3$           & $^\dagger$              & ---                     & $^\dagger$              \\
    VAGE$^*$         & $91.34\pm0.3$           & $^\dagger$             & $92.92\pm0.2$           & $^\dagger$              & ---                     & $^\dagger$              \\
    DeepWalk$^*$     & $90.71\pm0.6$           & $^\dagger$             & $83.10\pm0.5$           & $^\dagger$              & ---                     & $^\dagger$              \\
    Node2Vec$^*$     & $91.48\pm0.3$           & $^\dagger$             & $84.58\pm0.5$           & $^\dagger$              & ---                     & $^\dagger$              \\
    GAT$^*$          & $\textbf{94.73}\pm0.2$           & $91.27\pm0.4$          & $97.33\pm0.2$           & $95.37\pm0.3$           & $67.57\pm0.4$           & $62.32\pm0.5$           \\
    GraphSAGE$^*$    & $93.56\pm0.3$           & $91.09\pm0.3$          & $97.65\pm0.2$           & $\textbf{96.27}\pm0.2$           & $65.79\pm0.6$           & $60.13\pm0.6$           \\
    CTDNE            & $92.17\pm0.5$           & $^\dagger$             & $91.41\pm0.3$           & $^\dagger$              &     ---                 & $^\dagger$              \\
    Jodie            & $94.62\pm0.5$           & $\textbf{93.11}\pm0.4$          & $97.11\pm0.3$           & $94.36\pm1.1$           & $\textbf{\color{violet}85.20}\pm2.4$           & $\textbf{\color{violet}79.83}\pm2.5$           \\
    TGAT             & $\textbf{\color{violet}95.34}\pm0.1$           & $\textbf{\color{violet}93.99}\pm0.3$          & $\textbf{\color{violet}98.12}\pm0.2$           & $\textbf{\color{violet}96.62}\pm0.3$           & $70.02\pm0.6$           & $66.35\pm0.8$           \\
    DyRep            & $94.59\pm0.2$           & $92.05\pm0.3$          & $\textbf{97.98}\pm0.1$           & $95.68\pm0.2$           & $\textbf{83.52}\pm3.0$           & $\textbf{78.38}\pm4.0$  \\
    {\bf TGN-attn}   & $\textbf{\color{red}98.46}\pm0.1$  & $\textbf{\color{red}97.81}\pm0.1$ & $\textbf{\color{red}98.70}\pm0.1$  & $\textbf{\color{red}97.55}\pm0.1$  & $\textbf{\color{red}94.52}\pm0.5$  & $\textbf{\color{red}91.37}\pm1.1$           \\
    \bottomrule
  \end{tabular}\vspace{-5mm}
\end{table}


\subsection{Performance}

\begin{wraptable}{r}{6.5cm}\vspace{-4mm}
\caption{ROC AUC \% for the dynamic node classification.  $^*$Static graph method.}
  \label{tab:node-classification}
  \centering
  \begin{tabular}{lll}
    \toprule
                  & Wikipedia                 & Reddit                   \\
    \midrule
    GAE$^*$        & $74.85\pm0.6$           & $58.39\pm0.5$          \\
    VAGE$^*$       & $73.67\pm0.8$           & $57.98\pm0.6$          \\
    GAT$^*$        & $82.34\pm0.8$           & $\textbf{64.52}\pm0.5$          \\
    GraphSAGE$^*$  & $82.42\pm0.7$           & $61.24\pm0.6$          \\
    CTDNE          & $75.89\pm0.5$           & $59.43\pm0.6$          \\
    JODIE          & $\textbf{\color{violet}84.84}\pm1.2$           & $61.83\pm2.7$          \\
    TGAT           & $83.69\pm0.7$           & $\textbf{\color{violet}65.56}\pm0.7$          \\
    DyRep          & $\textbf{84.59}\pm2.2$           & $62.91\pm2.4$          \\
    {\bf TGN-attn} & $\textbf{\color{red}87.81}\pm0.3$  & $\textbf{\color{red}67.06}\pm0.9$ \\
    \bottomrule
  \end{tabular}
\vspace{-7mm}
\end{wraptable}

Table \ref{tab:edge-prediction} presents the results on future edge prediction. Our model clearly outperforms the baselines by a large margin in both transductive and inductive settings on all datasets. The gap is particularly large on the Twitter dataset, where we outperfom the second-best method (DyRep) by over 4\% and 10\% in the transductive and inductive case respectively. 
Table \ref{tab:node-classification} shows the results on dynamic node classification, where again our model obtains state-of-the-art results, with a large improvement over all other methods.

\begin{figure}\vspace{-5mm}
  \begin{subfigure}[t]{.465\textwidth}
    \centering
    \includegraphics[width=\textwidth]{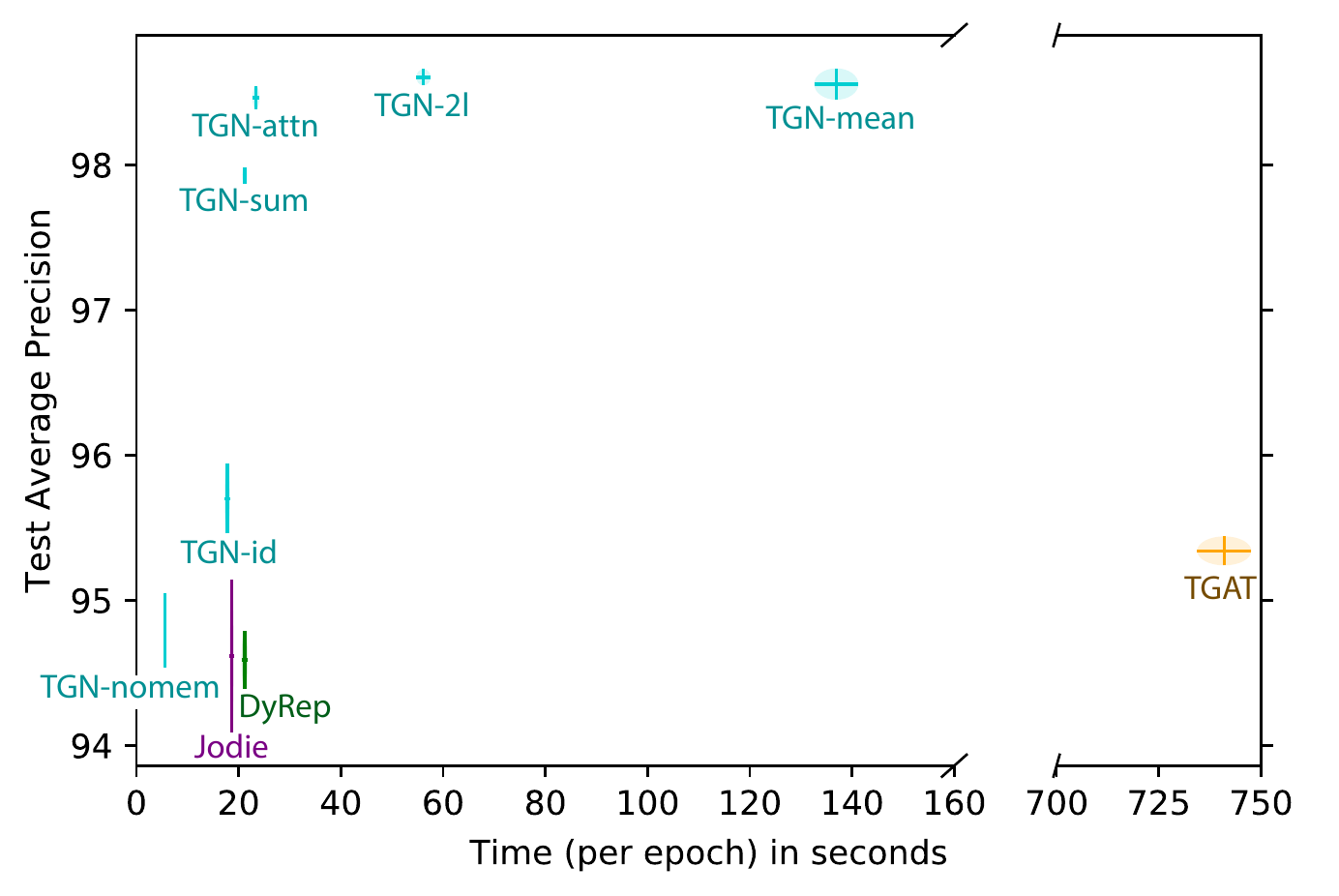}
    \caption{Tradeoff between accuracy (test average precision in \%) and speed (time per epoch in sec) of different models.}
    \label{fig:ablation}
  \end{subfigure}\hfill%
  \begin{subfigure}[t]{.45\textwidth}
    \centering
    \includegraphics[width=\textwidth]{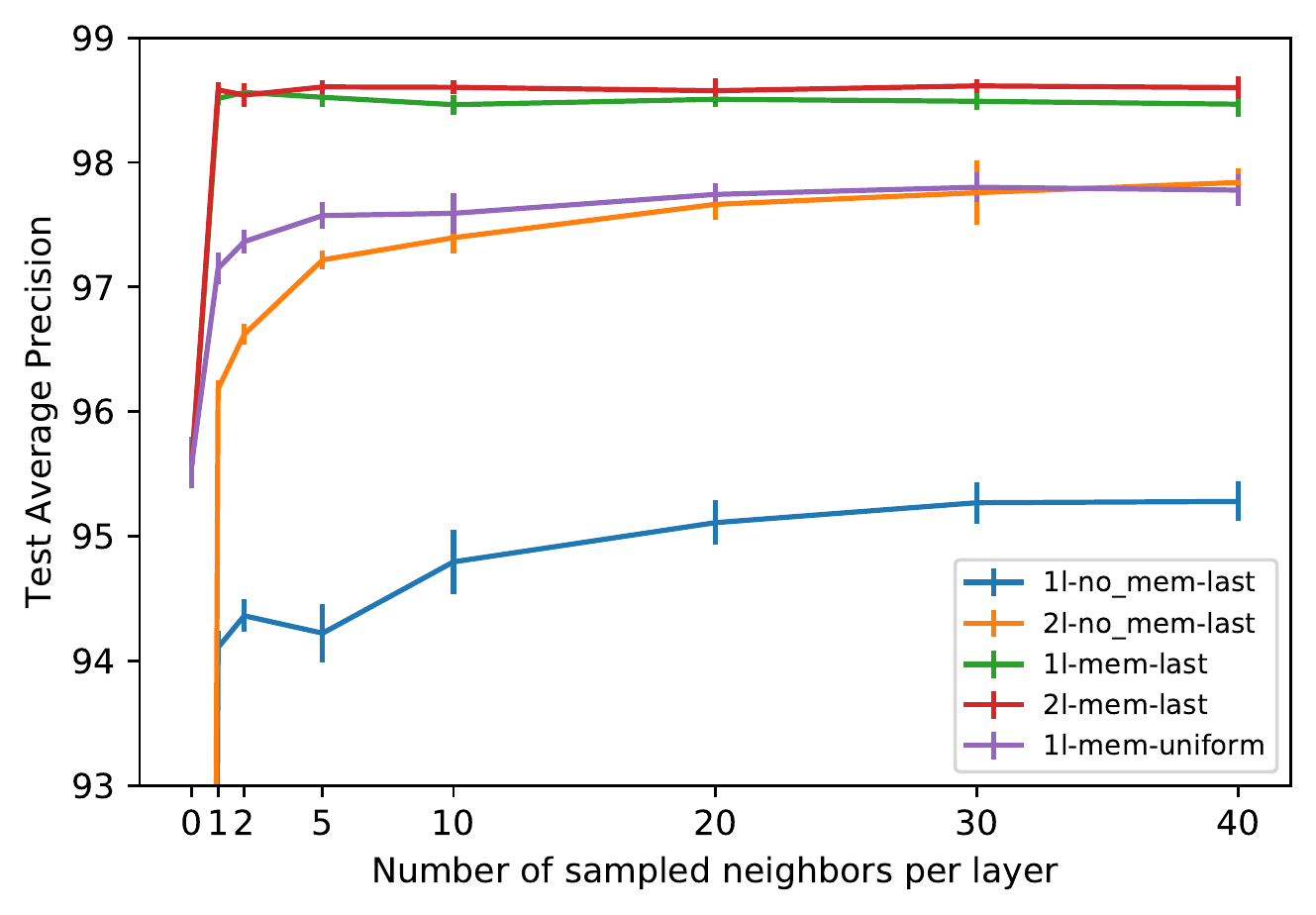}
    \caption{Performance of methods with different number of layers and with or without memory when sampling an increasing number of neighbors. {\em Last} and {\em uniform} refer to neighbor sampling strategy.}
    \label{fig:num_neighbors}
  \end{subfigure}%
  \caption{Ablation studies on the Wikipedia dataset for the transductive setting of the future edge prediction task. Means and standard deviations were computed over 10 runs. \vspace{-4mm}}
  \label{fig:ab}
  
\end{figure}

Due to the efficient parallel processing and the need for only one graph attention layer (see Section~\ref{sec:ablation_modules} for the ablation study on the number of layers), our model is up to $30\times$ faster than TGAT per epoch (Figure~\ref{fig:ablation}), while requiring a similar number of epochs to converge.

\subsection{Choice of Modules} \label{sec:ablation_modules}

We perform a detailed ablation study comparing different instances of our TGN framework, focusing on the speed vs accuracy tradeoff resulting from the choice of modules and their combination. The variants we experiment with are reported in Table~\ref{tab:models} and their results are depicted in Figure~\ref{fig:ablation}. 

\paragraph{Memory.} We compare a model that does not make use of a memory (TGN-no-mem), with a model which uses memory (TGN-attn) but is otherwise identical. While TGN-att is about $3\times$ slower, it has nearly 4\% higher precision than TGN-no-mem, confirming the importance of memory for learning on dynamic graphs, due to its ability to store long-term information about a node which is otherwise hard to capture. This finding is confirmed in Figure~\ref{fig:num_neighbors} where we compare different models when increasing the number of sampled neighbors: the models with memory consistently outperform the models without memory. Moreover, using the memory in conjunction with sampling the most recent neighbors reduces the number of neighbors needed to achieve the best performance when used.

\paragraph{Embedding Module.} Figure~\ref{fig:ablation} compare models with different embedding modules (TGN-id, TGN-time, TGN-attn, TGN-sum). The first interesting insight is that projecting the embedding in time seems to slightly hurt, as TGN-time underperforms TGN-id. Moreover, the ability to exploit the graph is crucial for performance:  we note that all graph-based projections (TGN-attn, TGN-sum) outperform the graph-less TGN-id model by a large margin, with TGN-attn being the top performer at the expense of being only slightly slower than the simpler TGN-sum. This indicates that the ability to obtain more recent information through the graph, and to select which neighbors are the most important are critical factors for the performance of the model.

\paragraph{Message Aggregator.} We compare two further models, one using the last message aggregator (TGN-attn) and another a mean aggregator (TGN-mean) but otherwise the same. While TGN-mean performs slightly better, it is more than $3\times$ slower.  

\paragraph{Number of layers.} While in TGAT having 2 layers is of fundamental importance for obtaining good performances (TGAT vs TGAT-1l has over 10\% difference in average precision), in TGN the presence of the memory makes it enough to use just one layer to obtain very high performance (TGN-attn vs TGN-2l). This is because when accessing the memory of the 1-hop neighbors, we are indirectly accessing information from hops further away. Moreover, being able to use only one layer of graph attention speeds up the model dramatically.

\section{Conclusion}
We introduce TGN,  a generic framework for learning on continuous-time dynamic graphs. We obtain state-of-the-art results on several tasks and datasets while being faster than previous methods. Detailed ablation studies show the importance of the memory and its related modules to store long-term information, as well as the importance of the graph-based embedding module to generate up-to-date node embeddings. We envision interesting applications of TGN in the fields of social sciences, recommender systems, and biological interaction networks, opening up a future research direction of exploring more advanced settings of our model and understanding the most appropriate domain-specific choices.

\bibliography{tgn_arxiv}

\appendix
\section{Appendix}

\subsection{Deletion Events} \label{sec:deletion_events}
The TGN frameworks also support edge and node deletions events.

In the case of an {\bf edge deletion} event $(i, j, t', t)$ where an edge between nodes $i$ and $j$ which was created at time $t'$ is deleted at time $t$, two messages can be computed for the source and target nodes that respectively started and received the interaction:
\begin{eqnarray}
    \mathbf{m}_i(t) = \mathrm{msg_{s'}}\left(\mathbf{s}_i(t^-), \mathbf{s}_j(t^-), \Delta t, \mathbf{e}_{ij}(t)\right), \quad\quad
    \mathbf{m}_j(t) = \mathrm{msg_{d'}}\left(\mathbf{s}_j(t^-), \mathbf{s}_i(t^-), \Delta t, \mathbf{e}_{ij}(t)\right)
\end{eqnarray}

In case of a {\bf node deletion} event, we simply remove the node (and its incoming and outgoing edges) from the temporal graph so that when computing other nodes embedding this node is not used during the temporal graph attention. Additionally, it would be possible to compute a message from the node's feature and memory and use it to update the memories of all its neighbors.

\subsection{TGN Training} \label{sec:pseudocode}
When parallelizing the training of TGN, it is important to mantain the temporal dependencies between interactions. If the events on each node were independent, we could treat them as a sequence and train an RNN on each node independently using Back Propagation Through Time (BTPP). However, the graph structure introduces dependencies between the events (the update of a node depends on the current memory of other nodes) which prevent us from processing nodes in parallel. Previous methods either process the interactions one at a time, or use the t-batch~\citep{10.1145/3292500.3330895} training algorithm, which however does not satisfy temporal consistency when aggregating from the graph as in the case of TGN (since the update does not only depend on the memory of the other node involved the interaction, but also on the neighbors of the two nodes).

\begin{figure}
    \centering
    \includegraphics[width=8.07cm]{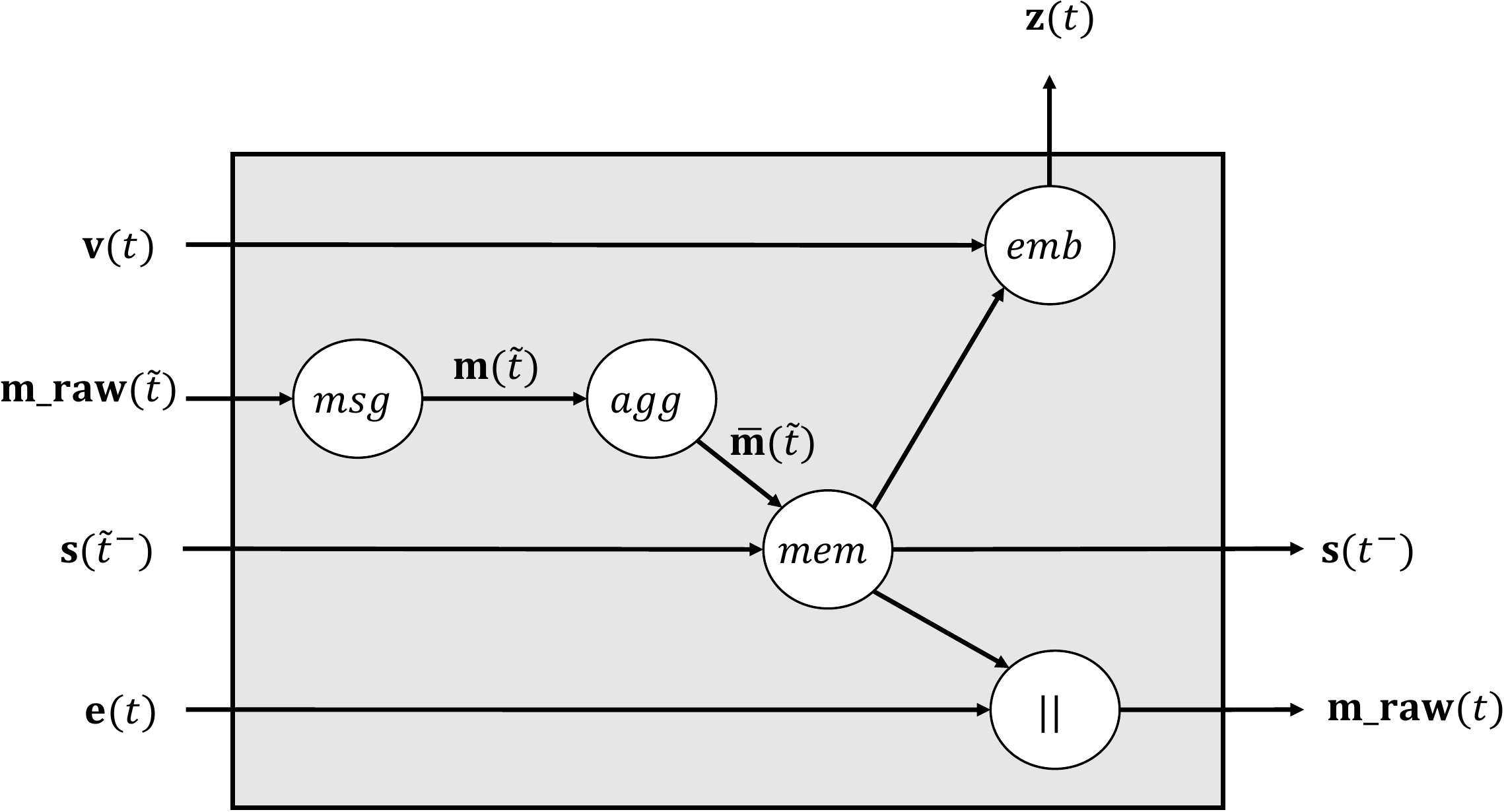}
    \caption{Schematic diagram of TGN. $\mathbf{m\_raw}(t)$ is the raw message generated by event $\mathbf{e}(t)$, $\tilde{t}$ is the instant of time of the last event involving each node, and $t^-$ the one immediately preceding $t$. }
    \label{fig:tgn_schematic_diagram}
\end{figure}

This issues motivate our training algorithm, which processes all interactions in batches following the chronological order. It stores the last message for each node in a message store, to process it before predicting the next interaction for the node. This allows the memory-related modules to receive a gradient. Algorithm \ref{algo:tgn-advanced-training} presents the pseudocode for TGN training, while Figure \ref{fig:tgn_schematic_diagram} shows a schematic diagram of TGN.

\begin{algorithm}
\SetAlgoLined
 $\mathbf{s} \leftarrow \mathbf{0} $    \tcp*{Initialize memory to zeros}
 $\mathbf{m\_raw} \leftarrow \{\} $    \tcp*{Initialize raw messages}
 
 \ForEach{batch $(\mathbf{i}, \mathbf{j}, \mathbf{e}, \mathbf{t}) \in$ training data}{
   $\mathbf{n} \leftarrow$ sample negatives ; \\
    $\mathbf{m} \leftarrow$ msg$(\mathbf{m\_raw}$)     \tcp*{Compute messages from raw features\footnotemark[1]} 
 
   $\mathbf{\bar{m}} \leftarrow$ agg($\mathbf{m}$)                                          \tcp*{Aggregate messages for the same nodes}
 
   $\mathbf{\hat{s}} \leftarrow \mathrm{mem}(\mathbf{\bar{m}}, \mathbf{s}$)              \tcp*{Get updated memory}
   
   $\mathbf{z_{i}}, \mathbf{z_{j}}, \mathbf{z_{n}} \leftarrow \mathrm{emb}_\mathbf{\hat{s}}(\mathbf{i}, \mathbf{t}), \mathrm{emb}_\mathbf{\hat{s}}(\mathbf{j}, \mathbf{t}), \mathrm{emb}_\mathbf{\hat{s}}(\mathbf{n}, \mathbf{t})$      \tcp*{Compute node embeddings\footnotemark[3]}
   $\mathbf{p_{pos}}$, $\mathbf{p_{neg}} \leftarrow \mathrm{dec}$($\mathbf{z_{i}}$, $\mathbf{z_{j}}$), $\mathrm{dec}$($\mathbf{z_{i}}$, $\mathbf{z_{n}}$)      \tcp*{Compute interactions probs}
   $l$ = BCE($\mathbf{p_{pos}}$, $\mathbf{p_{neg}}$)                        \tcp*{Compute BCE loss}
   $\mathbf{m\_raw_i}, \mathbf{m\_raw_j} \leftarrow (\mathbf{\hat{s}}_{\mathbf{i}}, \mathbf{\hat{s}}_{\mathbf{j}}, \mathbf{t}, \mathbf{e}), (\mathbf{\hat{s}}_{\mathbf{j}}, \mathbf{\hat{s}}_{\mathbf{i}}, \mathbf{t}, \mathbf{e})$             \label{raw_messages}                 \tcp*{Compute raw messages}
   $\mathbf{m\_raw} \leftarrow$ store\_raw\_messages($\mathbf{m\_raw}, \mathbf{m\_raw_i}, \mathbf{m\_raw_j}$)        \tcp*{Store raw messages}
   $\mathbf{s}_{\mathbf{i}}, \mathbf{s}_{\mathbf{j}} \leftarrow \mathbf{\hat{s}}_{\mathbf{i}}, \mathbf{\hat{s}}_{\mathbf{j}}$ \tcp*{Store updated memory for sources and destinations}
 }
 \caption{Training TGN}
 \label{algo:tgn-advanced-training}
\end{algorithm}
\footnotetext[1]{For the sake of clarity, we use the same message function for both sources and destination.}
\footnotetext[2]{We denote with $\mathrm{emb}_\mathbf{\hat{s}}$ an embedding layer that operates on the updated version of the memory $\mathbf{\hat{s}}$.}

\subsection{Datasets} \label{sec:datasets}
Reddit and Wikipedia are bipartite interaction graphs. 
In the Reddit dataset, users and sub-reddits are nodes, and an interaction occurs when a user writes a post to the sub-reddit. In the Wikipedia dataset, users and pages are nodes, and an interaction represents a user editing a page. 
In both aforementioned datasets, the interactions are represented by text features (of a post or page edit, respectively), and labels represent whether a user is banned. 
Both interactions and labels are time-stamped. 

The Twitter dataset is a non-bipartite graph released as part of the 2020 RecSys Challenge \citep{belli2020privacy}. 
Nodes are users and interactions are retweets. The features of an interaction are a BERT-based \citep{Wolf2019HuggingFacesTS} vector representation of the text of the retweet. 

Node features are not present in any of these datasets, and we therefore assign the same zero feature vector to all nodes. Moreover, While our framework is general and in section \ref{sec:core_modules} we showed how it can process any type of event, these three datasets only contain the edge creation (interaction) event type. Creating and evaluation of datasets with a wider variety of events is left as future work. 

The statistics of the three datasets are reported in table \ref{tab:dataset_statistics}.

    \begin{table}[h]
      \caption{Statistics of the datasets used in the experiments.}
      \label{tab:dataset_statistics}
      \centering
      \begin{tabular}{llll}
        \toprule
                                      & Wikipedia                 & Reddit                  & Twitter                 \\
        \midrule
        \# Nodes                       & 9,227                    & 11,000                   & 8,861                  \\
        \# Edges                       & 157,474                   & 672,447                 & 119,872                \\
        \# Edge features               & 172                       & 172                     & 768                    \\
        \# Edge features type          & LIWC                      & LIWC                    & BERT                   \\
        Timespan                       & 30 days                   & 30 days                 & 7 days                 \\
        Chronological Split            & 70\%-15\%-15\%            & 70\%-15\%-15\%          & 70\%-15\%-15\%         \\
        \# Nodes with dynamic labels   & 217                       & 366                     & --                     \\
        \bottomrule
      \end{tabular}
    \end{table}
 
  \paragraph*{Twitter Dataset Generation} To generate the Twitter dataset we started with the snapshot of the Recsys Challenge training data on 2020/09/06. We filtered the data to include only retweet edges (discarding other types of interactions) where the timestamp was present. This left approximately 10\% of the edges in the original dataset. We then filtered the retweet multi-graph (users can be connected by multiple retweets) to only include the largest connected component. Finally, we filtered the graph to only the top 5,000 nodes in-degree and the top 5,000 by out-degree, ending up with 8,861 nodes since some nodes were in both sets. 

\subsection{Additional Experimental Settings and Results} \label{sec:additional_exp_setting}

\paragraph{Hyperparameters}
For the all datasets, we use the Adam optimizer with a learning rate of $0.0001$, a batch size of $200$ for both training, validation and testing, and early stopping with a patience of 5. We sample an equal amount of negatives to the positive interactions, and use \textit{average precision} as reference metric. Additional hyperparameters used for both future edge prediction and dynamic node classification are reported in table \ref{tab:hyperparams}.
For all the graph embedding modules we use neighbors sampling \citep{hamilton2017representation} (i.e. only aggregate from $k$ neighbors) since it improves the efficiency of the model without losing in accuracy. In particular, the sampled edges are the $k$ most recent ones, rather than the traditional approach of sampling them uniformly, since we found it to perform much better (see Figure \ref{fig:neighbor_sampling}). All experiments and timings are conducted on an AWS p3.16xlarge machine and the results are averaged over 10 runs. The code will be made available for all our experiments to be reproduced.

  \begin{table}[h]
      \caption{Model Hyperparameters.}
      \label{tab:hyperparams}
      \centering
      \begin{tabular}{ll}
        \toprule
                                      & Value                 \\
        \midrule
        Memory Dimension            & 172                                  \\
        Node Embedding Dimension    & 100                                   \\
        Time Embedding Dimension    & 100                                   \\
        \# Attention Heads          & 2                        \\
        Dropout                     & 0.1                                \\
        \bottomrule
      \end{tabular}
    \end{table}
    
\paragraph{Baselines Results}
Our results for GAE~\citep{kipf2016variational}, VGAE~\citep{kipf2016variational}, DeepWalk~\citep{10.1145/2623330.2623732},  Node2Vec~\citep{10.1145/2939672.2939754}, GAT \citep{DBLP:conf/iclr/VelickovicCCRLB18} and GraphSAGE \citep{hamilton2017representation}, CTDNE~\citep{10.1145/3184558.3191526} and TGAT~\citep{Xu2020Inductive} are taken directly from the TGAT paper~\citep{Xu2020Inductive}. For Jodie~\citep{10.1145/3292500.3330895} and DyRep~\citep{trivedi2018dyrep}, in order to make the comparison as fair as possible, we implement our own version in PyTorch as a specific case of our tgn framework. For Jodie we simply use the time embedding module, while for DyRep we augment the messages with the result of a temporal graph attention performed on the destination's neighborhood. For both we use a vanilla RNN as the memory updater module.

\subsubsection{Neighbor Sampling: Uniform vs Most Recent}
\begin{figure}[h]
  \centering
  \includegraphics[width=.5\textwidth]{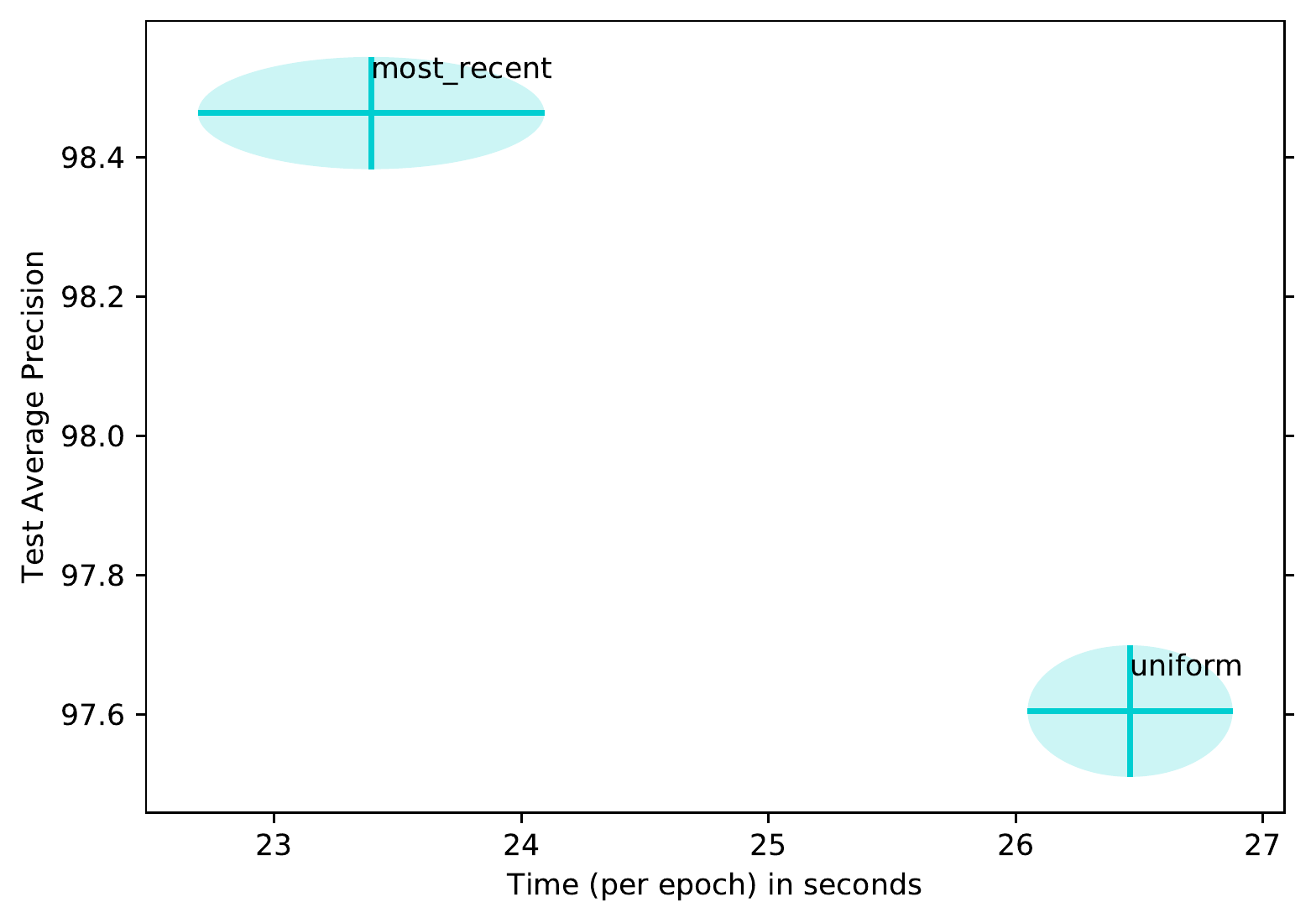}
  \caption{  \label{fig:neighbor_sampling} Comparison of two TGN-attn models using different neighbor sampling strategies (when sampling 10 neighbors). Sampling the most recent edges clearly outperforms uniform sampling. Means and standard deviations (visualized as ellipses) were computed over 10 runs. }
\end{figure}

When performing neighborhood sampling~\citep{graphsage} in static graphs, nodes are usually sampled uniformly. While this strategy is also possible for dynamic graphs, it turns out that the most recent edges are often the most informative. In Figure \ref{fig:neighbor_sampling} we compare two TGN-attn models (see Table \ref{tab:models}) with either uniform or most recent neighbor sampling, which shows that a model which samples the most recent edges obtains higher performances. 

\end{document}